\documentclass[smallextended]{svjour3}
\smartqed
\usepackage{graphicx}
\usepackage{mathptmx}
\usepackage{times}
\usepackage{graphicx}
\usepackage{latexsym}
\usepackage{amsmath, amssymb}
\usepackage{url}
\journalname{Annals of Operations Research}
\begin{document}
\title{Improved Branch-and-Bound for\\
Low Autocorrelation Binary Sequences}
\titlerunning{Improved Branch-and-Bound for LABS}
\author{S. D. Prestwich}
\institute{Department of Computer Science,
University College Cork, Ireland\\
Tel. +353 21 420 5911\\
Fax +353 +353 21 420 5367\\
\email{s.prestwich@cs.ucc.ie}}
\date{}
\maketitle
\begin{abstract}

The Low Autocorrelation Binary Sequence problem has applications in
telecommunications, is of theoretical interest to physicists, and has
inspired work by many optimisation researchers because of its
difficulty.  For many years it was considered unsuitable for solution
by metaheuristics because of its search space topology, but in recent
years metaheuristics have found long high-quality sequences.  However,
complete search has not progressed since a parallel branch-and-bound
method of 1996.  In this paper we find four ways of improving
branch-and-bound, leading to a tighter relaxation, faster convergence
to optimality and better scalability.  We also extend known optimality
results for skew-symmetric sequences from length 73 to 89.
\end{abstract}

\newpage

\section{The LABS problem}

Consider a binary sequence $S=(s_1,\ldots,s_N)$ where each $s_i \in \{1,-1\}$.
The {\it off-peak autocorrelations\/} of $S$ are defined as
\begin{eqnarray}
C_k(S) &=& \sum_{i=1}^{N-k}s_is_{i+k} \;\;\; (k=1 \ldots N-1)
\label{ckcon}
\end{eqnarray}
and the {\it energy\/} of $S$ as
\begin{eqnarray}
E(S) &=& \sum_{k=1}^{N-1}C_k^2(S)
\label{econ}
\end{eqnarray}
The {\it low-autocorrelation binary sequence\/} (LABS) problem is to
assign values to the $s_i$ such that $E(S)$ is minimized.  A common
measure of sequence quality is the {\it merit factor\/}
$F(S)=N^2/2E(S)$.  Theoretical considerations \cite{Gol} give an upper
bound on $F(S)$ of approximately 12.32 as $N \rightarrow \infty$, and
empirical curve fitting on known optimal sequences \cite{Mer} yields
an estimate of $F \approx 9.3$ for large $N$.  This problem has many
practical applications in communications engineering, and is of
theoretical interest to physicists because it models 1-dimensional
systems of Ising-spins.  LABS has also generated interest among
researchers from other fields who are interested in hard optimisation
problems, and is problem number 5 in the CSPLib benchmark library
\cite{GenWal}, a web-based collection of constraint problems (though
it has no constraints).  Analytical methods have been used to
construct optimal sequences for certain values of $N$ (see
\cite{MerBes} for example) but for the general case search is
necessary.  Two possibilities are {\it complete\/} and {\it
  incomplete\/} search.

Complete search usually involves the enumeration of possibilities by
backtracking.  \cite{Gol} used exhaustive enumeration to find optimal
sequences for $N \le 32$.  \cite{Rei} used a complete enumeration
method with an accelerated function evaluation to compute exact
solutions for $N \le 39$.  \cite{Mer} enumerated optimal sequences for
$N \le 48$ using complete search augmented with two techniques to
reduce the size of the search tree: {\it branch-and-bound\/} and {\it
  symmetry breaking\/}.  Symmetry breaking, sometimes called {\it
  symmetry exclusion\/}, exploits the fact that sequences occur in
equivalence classes of size 8 (see Section \ref{sym}).  However, even
with these enhancements, complete search is unlikely to scale up to
large sequences, and it is conjectured that for $N>100$ progress will
be made through mathematical insight rather than computer power
\cite{Mer}.  The only other complete approach we know of is a recent
Quadratic Programming model \cite{Kra} which turned out to be much
slower than Mertens' method.  It was only used to solve instances up
to $N=30$ by which time it took tens of thousands of seconds, whereas
Mertens' method took a few seconds on an older machine.

When complete search becomes impractical it is common to resort to
incomplete methods such as simulated annealing, evolutionary
algorithms, neural networks, ant colonies or greedy algorithms, which
are often able to solve much larger instances.  Unfortunately they
perform quite poorly on some problems, and finding optimal LABS
solutions seemed for several years to be an example.  The cause was
considered to be the search space, whose cost function $E$ has a very
irregular structure with isolated minima \cite{Ber}.  However, more
recent approaches show that metaheuristics can find optimal sequences
efficiently.  Examples of metaheuristic methods applied to LABS are
simulated annealing \cite{Ber,Gol}, evolutionary search
\cite{DegEtc,MilEtc,Muh,Rei,Wan}, TABU search
\cite{DotHen,HalEtc,HulSok}, memetic algorithms \cite{GalEtc}, local
search \cite{BeeEtc,BrgEtc}, and local search hybridised with
relaxation \cite{Pre}.

In this paper we find several ways of improving branch-and-bound for
LABS, show that the new algorithm has improved scalability, and
establish new optimality results for skew-symmetric sequences.  The
method is described in Section \ref{method} and results are presented
in Section \ref{results}.

\section{Improved branch-and-bound for LABS} \label{method}

First we describe Mertens' branch-and-bound method \cite{Mer}, then
introduce our improvements.

\subsection{Mertens' method}

In order to minimize the minimum energy
\[
E_{\mbox{\footnotesize min}}=\mbox{min}\left( \sum_{k=1}^{N-1}C_k^2 \right)
\]
of a partial sequence $A$, the relaxation
\[
E^{*}_{\mbox{\footnotesize min}}=\sum_{k=1}^{N-1}\mbox{min}(C_k^2)
\]
can be used as a lower bound $E^*_{\mbox{\footnotesize min}} \le
E_{\mbox{\footnotesize min}}$.  Because $E^{*}_{\mbox{\footnotesize
    min}}$ is still expensive to calculate, Mertens' method uses a
cheap lower bound $E_b \le E^{*}_{\mbox{\footnotesize min}}$ based on
an arbitrary completion of the current partial sequence.  Define a
product $s_i s_j$ to be {\it computable\/} if $s_i$ and $s_j$ are both
assigned in the current partial sequence $A$.  Let $t_k$ be the sum of
its computable products and $f_k$ the number of its uncomputable
products.  A lower bound $l_k$ for each $C_k$ is calculated by finding
the energy of the completed sequence then subtracting $2f_k$, because
negating an element cannot reduce $C_k$ by more than 2.  On finding a
sequence with energy $E$ the search can proceed with upper bound $E-4$
because it is known that these sequences have energies differing by
multiples of 4 \cite{MilEtc}.

A refinement exploits the fact that the sum of an odd number of $\pm
1$ values has absolute value at least 1.  $l_k$ is refined to
$\max(l_k,b_k)$ where $b_k=(N-k)\,\mbox{mod}\,2$.  Then
\[
E_{b} = \sum_{k=1}^{N-1} l^2_k
\]
is a lower bound for $E^{*}_{\mbox{\footnotesize min}}$.

Symmetry occurs in LABS because the energy of a sequence is unaffected
if the sequence is reversed, if all its values are negated, or if
oddly numbered values are negated: combining these three operations in
all possible ways gives 8 equivalent sequences.  If we can avoid
exploring more than 1 sequence from each class we might reduce search
effort by a factor approaching 8, and Mertens' method achieves this by
fixing the values of several of the outermost variables.  The
branching heuristic is chosen to facilitate symmetry breaking:
variables are assigned outermost first, that is $s_1, s_N, s_2,
s_{N-1}, s_3, s_{N-2},\ldots$ (actually they are treated as pairs
$(s_1,s_N)$, $(s_2,s_{N-1})$, $\ldots$).

\subsection{Avoiding the use of an arbitrary sequence}

In Mertens' method the value of $l_k$ depends on the arbitrary
completion of the sequence, and its greatest possible value occurs
when arbitrarily completing the sequence transforms each uncomputable
product to $-1$ in which case $l_k=\max(b_k,|t_k|-f_k)$.  But we can
always achieve this value by reasoning as follows.  To the known sum
$t_k$ of computable products we add $f_k$ uncomputable products.  If
$t_k>0$ then the worst case is that each uncomputable product is $-1$,
while if $t_k<0$ the worst case is that each uncomputable product is 1
(if $t_k=0$ then $l_k=b_k$), so we can use $l_k=\max(b_k,|t_k|-f_k)$.
This idea was previously used in a hybrid local search algorithm
\cite{Pre}.

\subsection{Exploiting cancellations}

$f_k$ is the number of uncomputable products, but we can ignore some
of these products because they can be predicted to cancel each other
out.  Suppose we have two products $s_ps_q$ and $s_qs_r$ where
$s_p,s_r$ have been assigned different values, but $s_q$ has not yet
been assigned.  Whichever value $s_q$ takes the two products will have
different values, so we can subtract 2 from $f_k$ without knowing the
value of $s_q$ (of course we must remember not to repeat this
subtraction later in the search when $s_q$ is assigned a value).  We
shall refer to this as a {\it cancellation\/}.

Cancellations occur if we use the same branching heuristic as Mertens.
Suppose we have just assigned $s_i$ and we compute some $l_k$, where
$i \le \lfloor N/2 \rfloor$, $i+2k \le N$, $s_{i+k}$ is unassigned,
$s_{i+2k}$ is assigned, and $s_i \neq s_{i+2k}$.  Then we have a
cancellation between products $s_is_{i+k}$ and $s_{i+k}s_{i+2k}$.
Similarly if $i \ge \lceil N/2 \rceil$, $i-2k \ge 1$, $s_{i-k}$ is
unassigned, $s_{i-2k}$ is assigned, and $s_i \neq s_{i-2k}$, then we
have a cancellation between products $s_is_{i-k}$ and
$s_{i-k}s_{i-2k}$.  If we instead ordered variables ($s_1, s_2, s_3,
\ldots$) then no cancellations would occur, so Mertens' branching
heuristic turns out to be ideal for cancellations as well as for
symmetry breaking.

\subsection{Exploiting reinforcements}

We can also increase the value of $b_k$ in some cases.  Again consider
two uncomputable products $s_ps_q$ and $s_qs_r$ where $s_p,s_r$ have
been assigned values but $s_q$ has not.  This time suppose $s_p,s_r$
take the {\it same\/} value so that no cancellation occurs.  Then the
two uncomputable products will sum to either 2 or $-2$.  We shall
refer to this as a {\it reinforcement\/}.  Again if we use Mertens'
branching heuristic reinforcements will often occur, and we exploit
them as follows.  If all uncomputable products occur either in
cancellation or reinforcement pairs then $f_k$ is an even number.  Now
suppose that $t_k$ is also an even number, and that
$t_k\,\mbox{mod}\,4 \neq f_k\,\mbox{mod}\,4$.  Then $t_k+f_k$ is even
but must be of the form $4i+2$ for some integer $i \in \mathbb{Z}$ so
we can set $b_k=2$.

\subsection{Value ordering}

Our final improvement does not tighten the relaxation but leads to
faster convergence.  Mertens' method presumably assigned each variable
first to 1 then to $-1$ or vice-versa, because this is standard
practice and no special value ordering was mentioned in \cite{Mer}.
In experiments we found that almost any other value ordering,
including a randomised ordering, led more quickly to lower-energy
sequences.  We found best results by basing the value ordering on a
known large sequence with low energy, as follows.

For each variable we choose a fixed value that will always be tried
first during search: we shall refer to the vector of these values as a
{\it template\/}.  The template is based on a low-energy sequence
found by local
search\footnote{http://www.comp.nus.edu.sg/\~{}stevenha/viz/results\_labs.html}
so this is a simple way of exploiting local search results in
branch-and-bound.  To construct a template, for odd $N$ we take the
middle $N$ values from the sequence
\begin{quote}
12112111211222B2221111111112224542
\end{quote}
which has length 67, energy 241 and merit factor 9.31, while for even $N$
we take the middle $N$ values from the sequence
\begin{quote}
11111111141147232123251412112221212
\end{quote}
which has length 68, energy 250 and merit factor 9.25.  These two
sequences were chosen because they have high merit factors and are
longer than any sequences we need in this paper.  They are shown in
{\it run-length notation\/} in which each number indicates the number
of consecutive elements with the same value.  For example the sequence
($1$, $1$, $-1$, $1$, $-1$, $-1$, $-1$, $-1$, $1$) would be written
21141: whether the sequence begins with $1$ or $-1$ is irrelevant
because of symmetry.  For runs of length greater than 9 upper-case
letters are used: A=10, B=11 and so on.

\begin{figure}
\begin{center}
\includegraphics[scale=0.7]{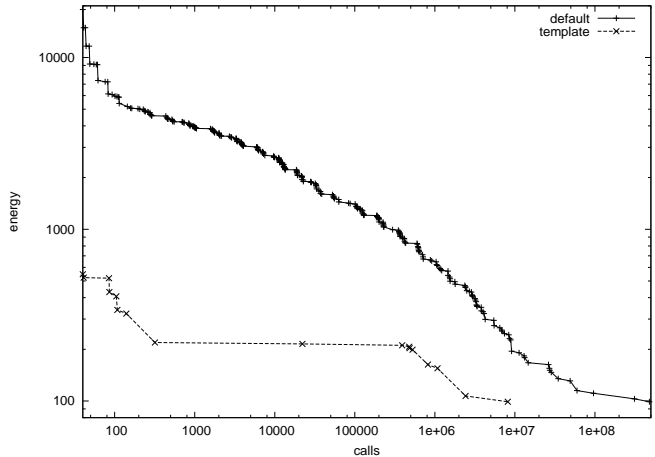}
\end{center}
\caption{Effect of a template on convergence}
\label{conv}
\end{figure}

The motivation behind this idea is that each correlation $C_k$ for the
new sequence takes all its terms from $C_k$ in the larger sequence.
While this does not guarantee optimality it should lead to a low
initial energy.  In experiments this does indeed occur, and using a
template greatly speeds up convergence to optimality.  For example the
graph in Figure \ref{conv} shows the runs for $N=39$ with and without
the use of a template.  The use of a template results in much earlier
low-energy sequences.  It also results in far fewer distinct energies,
which might aid future parallelisation by reducing communication
between processes: 307 energies without a template and 17 with.  The
effect on runtime is significant in many cases.  For example with
$N=39$ the method without a template (but with all other improvements)
takes 920 seconds to find an optimal sequence and a further 288
seconds to prove optimality, whereas with a template it takes 15
seconds to find an optimal sequence and a further 816 seconds to prove
optimality.

\subsection{Symmetry} \label{sym}

We break almost all symmetries in a similar way to Mertens, but taking
the template into account via a standard technique from Constraint
Programming.  If we were using a constraint model we could break all
symmetries by adding 7 {\it lex-leader constraints\/} \cite{CraEtc} to
ensure that any sequence is the lexicographically-least in its class.
For example to exploit the symmetry that results from reversing a
sequence and negating its values we would add a constraint
\[
\langle s_1, s_2, s_3, \ldots \rangle \preceq_{\mbox{\footnotesize lex}}
\langle -s_N, -s_{N-1}, -s_{N-2}, \ldots \rangle
\]
However, it is known that symmetry breaking can have a deleterious
effect on search if it conflicts with the search heuristics: that is,
if the excluded solutions are those that would have been found
earliest without symmetry breaking.  To avoid this conflict we ensure
that the template is the lexicographically-least among all possible
variable assignments, by adjusting the lex-leaders.  For example the
above lex-leader becomes
\[
\langle s'_1, s'_2, s'_3, \ldots \rangle \preceq_{\mbox{\footnotesize lex}}
\langle -s'_N, -s'_{N-1}, -s'_{N-2}, \ldots \rangle
\]
where
\[
s'_i= \left\{
\begin{array}{rl}
s_i \;\mbox{if}\; t_i=1\\
-s_i \;\mbox{if}\; t_i=0
\end{array}
\right.
\]
and $t_i$ is template value $i$ ($i=1 \ldots N$).  To reduce runtime
overhead we do not use the lex-leaders at every search tree node.
Instead we check that they are satisfied only at even-numbered depths
down to a depth of $N/2$, which is sufficient to break most symmetry
with low overhead.

\subsection{Skew-symmetry}

We can adapt our method to find only {\it skew-symmetric\/} sequences,
which is the most common {\it sieve\/} for restricting search to a
useful subset of all sequences \cite{Gol}.  The skew-symmetric
sequences have odd length with $N=2n-1$ for some $n$, and satisfy
\[
s_{n+i}=(-1)^ i s_{n-i} \;\;\; (i = 1 \ldots n-1)
\]
This roughly halves the number of independent variables in the
problem, which greatly reduces the search space.  Such sequences often
have good merit factors because $C_k=0$ for all odd $k$.  (Note that
skew-symmetry is a property of a single sequence, and should not be
confused with the 8-fold symmetry between sequences described above.)
Optimal skew-symmetric sequences have been enumerated using
branch-and-bound for $N \le 71$ by \cite{DegEtc} and for $N \le 73$ by
\cite{Rei}, and good solutions for larger $N$ have been found using
metaheuristics \cite{BeeEtc,DegEtc,GolHar,MilEtc,Muh,Pre,Rei,Wan}.

To adapt our branch-and-bound method to skew-symmetric sequences we
need three modifications.  Firstly we ensure that no assignment
violates skew-symmetry.  Secondly on finding a sequence with energy
$E$ we can use a new upper bound $E-8$.  Thirdly we need a longer
template that is also skew-symmetric, and we use a known sequence with
$N=119$:
\begin{quote}
11331111311332321211561311512
\end{quote}
(only the first 60 values are represented here as the rest can be
deduced by skew-symmetry) which has energy 835 and merit factor 8.48.

\section{Results and conclusions} \label{results}

Mertens tested the scalability of the branch-and-bound method by
counting the number of recursive calls needed to find an optimal
sequence and prove it optimal, using results for $N=15 \ldots 44$ then
curve-fitting (H. Bauke, personal communication) and found that it
needed $O(1.85^N)$ calls.  Performing the same experiment we find
improved scalability of $O(1.74^N)$ calls, or $O(1.80^N)$ seconds of
wall clock time, using a C implementation of our method executed on a
2.8 GHz Pentium 4 with 512 MB RAM.  We hope to use this method to find
new optimal sequences in the future, but to do this we need a parallel
implementation: since publishing \cite{Mer} Mertens and Bauke have
found provably optimal sequences up to $N=60$ using a cluster of 160
processors.  By extrapolation we do not expect a speedup of 160 to
occur until $N \approx 83$ so for the present parallelism trumps our
improvements.  But our new method should give good results when
implemented on a cluster and we hope to try this in future work.

However, we can use our method to find new results for skew-symmetric
sequences, which to the best of our knowledge have not been attacked
using highly parallel hardware.  In experiments we found no new
skew-symmetric sequences, but confirmed the optimality of several
published sequences previously found by metaheuristics.  Table
\ref{skewres} shows the merit factors and execution times.  Results
for $N \le 71$ can be found in \cite{DegEtc}, and because \cite{Rei}
may not be easy to obtain we mention that the optimal merit factor for
skew-symmetric sequences of length $N=73$ is 7.66.  The merit factor
8.25 for $N=75$ is optimal though \cite{BeeEtc} incorrectly gives it
as 9.25.

\begin{table}
\begin{center}
\begin{tabular}{rrr}
\hline\noalign{\smallskip}
$N$ & $F$ & seconds\\
\hline\noalign{\smallskip}
75 & 8.25 & 3,655\\
77 & 8.28 & 9,140\\
79 & 7.67 & 17,889\\
81 & 8.20 & 28,321\\
83 & 9.14 & 35,666\\
85 & 8.17 & 74,994\\
87 & 8.39 & 143,147\\
89 & 8.18 & 285,326\\
\hline\noalign{\smallskip}
\end{tabular}
\end{center}
\caption{New optimality results for skew-symmetric sequences}
\label{skewres}
\end{table}

There might be further possible improvements to the LABS
branch-and-bound algorithm.  If we expand the energy expression we
obtain a quartic polynomial which in principle allows more
cancellations.  Consider terms $s_a s_b s_c s_d$ and $s_a s_b s_c
s_e$.  If $s_a,s_b,s_d,s_e$ are assigned and if $s_a s_b s_d \neq s_a
s_b s_e$ then the two terms cancel out whatever the value of $s_c$.
Furthermore, if only $s_c,s_d,s_e$ are assigned and $s_c s_d \neq s_c
s_e$ the terms cancel out whatever the values of $s_a,s_b$.  And if
only $s_d,s_e$ are assigned and $s_d \neq s_e$ then the two terms
cancel out whatever the values of $s_a,s_b,s_c$.  The difficulty lies
in exploiting these additional cancellations efficiently, which is an
interesting possibility for future research.

\end{document}